\documentclass[twocolumn,final]{svjour3}          
\smartqed  
\usepackage[inline]{enumitem}
\usepackage{graphicx}
\usepackage{color}
\usepackage{amssymb}
\usepackage{booktabs}
\usepackage{amsmath,bm}
\usepackage{multirow}
\usepackage[colorlinks,linkcolor=red,anchorcolor=blue,citecolor=green,CJKbookmarks=True]{hyperref}
\usepackage[numbers]{natbib} 
\usepackage[linesnumbered,ruled,vlined]{algorithm2e}
\journalname{International Journal of Computer Vision}
\begin{document}
%
\title{Cooperative Semantic Segmentation and Image Restoration in Adverse Environmental Conditions}
\titlerunning{SR-Net}
%

\author{Weihao Xia \and Zhanglin Cheng \and Yujiu Yang \and Jing-Hao Xue}
\authorrunning{Weihao Xia et al.} 

\institute{Weihao Xia\at
Tsinghua University\\
\email{xiawh3@outlook.com}
\and
Zhanglin Cheng\at
Shenzhen Institutes of Advanced Technology (SIAT), Chinese Academy of Science\\
\email{zl.cheng@siat.ac.cn}
\and
Yujiu Yang\at
Tsinghua Shenzhen International Graduate School\\
\email{yang.yujiu@sz.tsinghua.edu.cn}
\and
Jing-Hao Xue\at
Department of Statistical Science, University College London\\
\email{jinghao.xue@ucl.ac.uk}
}

\date{Received: date / Accepted: date}

\maketitle

\begin{abstract}
Most state-of-the-art semantic segmentation approaches only achieve high accuracy in good conditions. 
In practically-common but less-discussed adverse environmental conditions, their performance can decrease enormously. 
Existing studies usually cast the handling of segmentation in adverse conditions as a separate post-processing step after signal restoration, making the segmentation performance largely depend on the quality of restoration. 
In this paper, we propose a novel deep-learning framework to tackle semantic segmentation and image restoration in adverse environmental conditions in a holistic manner. 
The proposed approach contains two components: \textit{Semantically-Guided Adaptation}, which exploits semantic information from degraded images to refine the segmentation; and \textit{Exemplar-Guided Synthesis}, which restores images from semantic label maps given degraded exemplars as the guidance. 
Our method cooperatively leverages the complementarity and interdependence of low-level restoration and high-level segmentation in adverse environmental conditions. 
Extensive experiments on various datasets demonstrate that our approach can not only improve the accuracy of semantic segmentation with degradation cues, but also boost the perceptual quality and structural similarity of image restoration with semantic guidance.
\end{abstract}

\section{Introduction}
\label{sec:intro}

\begin{figure}[t]
  \centering
  \centerline {\includegraphics[width=\linewidth]{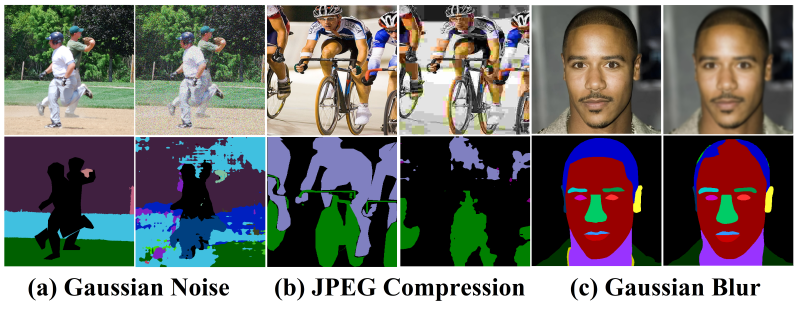}}
  \caption{The illustration of semantic segmentation in three adverse environmental conditions: (a) Gaussian noise; (b) JPEG compression; and (c) Gaussian blur. In each of panels (a)-(c), the left-hand column is for the normal condition, while the right-hand column is for the adverse conditions.}
  \label{fig:motivation}
\end{figure}

Reliable visual sensing and understanding algorithms are crucial for emerging applications, such as UAVs, search and rescue robots, autonomous driving and security surveillance. 
However, since most current vision systems are designed to perform under normal conditions, the performances of computer vision based sensing and understanding of outdoor environments will be largely jeopardized by various challenging conditions in unconstrained and dynamic degraded environments, e.g., moving platforms, bad weathers, and poor illumination; or more technically specifically, noise, compression and blur, as illustrated in Figure~\ref{fig:motivation}.
Therefore, it is highly desirable to study how to cope with such challenging visual conditions for achieving robust visual sensing and understanding in the wild.

Conventionally, low-level image processing tasks, e.g., image restoration, and high-level visual understanding problems, e.g., semantic segmentation, are separately tackled by different frameworks.
Most existing studies handle the challenging conditions as an independent step of image restoration, and then feed the restored data for subsequent segmentation. The performance of segmentation will thus largely depend on the quality of restoration.
It remains questionable whether restoration-based approaches would actually improve the visual understanding performance, as the restoration step may be suboptimal for the ultimate target task and may also bring in misleading information and lethal artifacts.

Recent reports~\citep{alBahar2019guided,tang2019multichannel,   wang2018pix2pixHD} have revealed the connection between low-level and high-level visual tasks. 
For example, \citet{wang2018pix2pixHD} propose to synthesize a photo-realistic image from a given semantic label; \citet{ma2019exemplar} transfer the style of the target exemplar to the source image, while preserving semantic consistency; \citet{alBahar2019guided} aim to generate a high-resolution depth map given a low resolution depth map with the guidance of a high resolution RGB image.
Inspired by their work, we present a novel deep-learning framework for semantic segmentation and image restoration under adverse environmental conditions. Our strategy is to make the low-level image restoration and high-level semantic segmentation cooperate with each other, by leveraging the complementary information and inherent interdependence of the two seemingly unrelated tasks in a holistic way. 

Our proposed method contains the following two components: 
\begin{enumerate*}[label=(\alph*)] \item \textit{Semantically-Guided Adaptation}, which exploits semantic information from degraded images to refine the segmentation; and \item \textit{Exemplar-Guided Synthesis}, which synthesizes restored images from semantic label maps given degraded exemplars.\end{enumerate*}
These two components are implemented by their corresponding models:
\begin{enumerate*}[label=(\alph*)] \item the \textit{Refinement Network}, and \item the \textit{Restoration Network}.\end{enumerate*}
The refinement network takes the degraded image and its corresponding semantic map as input, and is trained to produce refined segmentation, by exploiting the intrinsic connection between raw images and semantic information.
The restoration network restores the degraded image with the complementary semantic information from the obtained segmentation result.

We systematically investigate the mutual influence between the low-level image restoration and high-level semantic segmentation tasks through extensive experiments on various datasets: Cityscapes~\citep{Cordts2016Cityscapes}, PASCAL VOC 2012~\citep{Everingham2010VOC}, COCO-Stuff~\citep{Caesar2018COCO-Stuff}, ADE20K~\citep{zhou2017scene}, and  CelebAMask-HQ~\citep{CelebAMask-HQ}. 
The experiments demonstrate that our approach can not only improve the accuracy of high-level vision tasks with degradation cues, but also boost the perceptual quality and structural similarity of restored images with semantic guidance.

The paper is structured as follows. Section~\ref{sec:related work} presents the related work. Section~\ref{sec:method} is devoted to our method for semantically-guided adaptation and exemplar-guided synthesis in Section~\ref{sec:SGAda} and Section~\ref{sec:EGSyn}, respectively. Section~\ref{sec:experiment} presents our experimental results and Section~\ref{sec:conclusion} concludes this paper.
Our code including data generation scripts will be publicly available at~\href{github.com/xiaweihao/SR-Net}{our website}.

\begin{figure*}[t]
  \centering
  \centerline {\includegraphics[width=\textwidth]{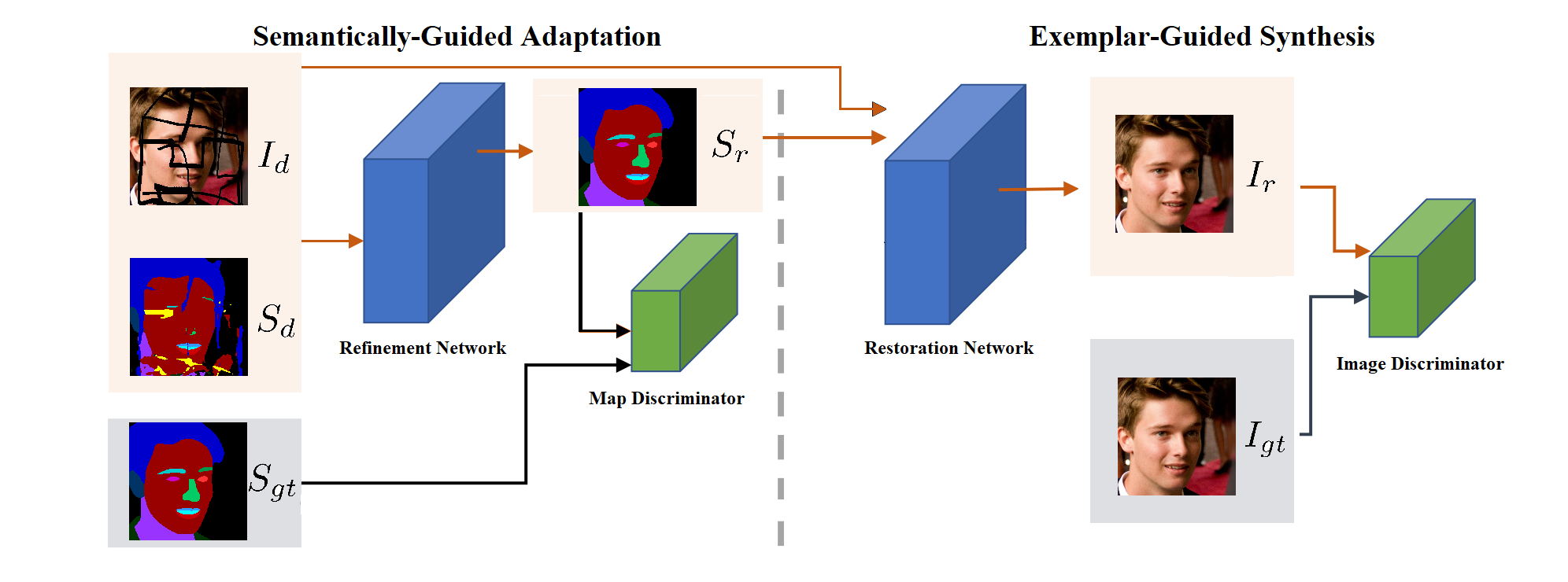}}
  \caption{Overview of the proposed architecture. Semantically-Guided Adaptation exploits and leverages semantic information from degraded images then help to refine the segmentation, and Exemplar-Guided Synthesis synthesizes the restored images from semantic label maps given specific degraded exemplars.}
  \label{fig:overview}
\end{figure*}

\section{Related Work}
\label{sec:related work}
\subsection{Semantic Segmentation in Adverse Environmental Conditions}
Most semantic segmentation algorithms focus on achieving high scores at well-known benchmarks, caring little about the robustness of their methods in adverse environmental conditions. In fact, there are only a few approaches which can deal with certain disturbances caused by diverse weather.

For instance,~\citet{SakaridisDG18} improve the segmentation accuracy in fog, by adding simulated fog to real-world images and training with these synthetic data.
\citet{Porav2019deraining} derain rainy images by means of a neural network, and their method achieves better performance when a further segmentation network is applied on the derained images.
\citet{BijelicMRKD18} introduce the Robust Learning Technique, by which an image classifier becomes more robust against unknown disturbances not occurring in the training set. They randomly choose one of the input channels and replace the corresponding input data by an arbitrarily selected sensor data of a different scene.
\citet{Valada2017adverse} select RGB, depth and EVI (Enhanced Vegetation Index) as the modalities to build expert networks for segmenting images in adverse environments.
\citet{Pfeuffer2019fusion} use a multiple-sensor setup, i.e., disturbed input images and corresponding depth images observed by the lidar sensor, to increase the robustness of semantic labeling approaches.

\subsection{Image Restoration for Multiple Degradations}
Most existing image restoration methods are designed for specific subtasks like image denoising of certain noise, lacking scalability in learning a single model for multiple degradations. Some recent studies aim to learn a single model to effectively handle multiple and even spatially-variant degradations.

\citet{ZhangZCM017DnCNN} introduce batch normalization into a single DnCNN model to jointly handle several image restoration tasks.
\citet{Mao2016Restoration} propose a 30-layer convolutional auto-encoder network named RED by introducing symmetric skip connections for image denoising and image super-resolution.
To add restricted long-term memory motivated by neocortical circuits,~\citet{TaiYLX17MemNet}  propose a very deep persistent memory network called MemNet, which introduces a memory block to explicitly mine persistent memory through an adaptive learning process. MemNet demonstrates unanimous superiority on three image restoration tasks, i.e., image denosing, super-resolution and JPEG de-blocking.
\citet{Zhang2018SRMD} learn a single convolutional super-resolution network SRMD with high scalability for multiple degradations. The proposed super-resolver takes both LR image and its degradation maps as input.~\citet{Liu2018NLRN} propose a non-local recurrent network (NLRN) as the first attempt to incorporate non-local operations into a recurrent neural network (RNN) and introduce the correlation propagation, which achieves superior results to state-of-the-art methods with many fewer parameters on image denoising and super-resolution tasks.
\citet{Ulyanov2018Prior} propose a non-trained method that using a randomly-initialized neural network as the handcrafted prior in standard inverse problems such as denoising, super-resolution, and inpainting.~\citet{Yu2018RLRestore} adopt an agent to select a toolchain to progressively restore an image corrupted by complex and mixed distortions by deep reinforcement learning.

\section{Methodology}
\label{sec:method}
\subsection{Overview of the Proposed Approach}
We propose a new deep-learning framework for cooperative semantic segmentation and image restoration in adverse environmental conditions. 
As illustrated in Figure~\ref{fig:overview},
given the semantic segmentation results obtained by methods optimized for normal conditions but the input images sensed in unconstrained and dynamic degraded environments, we refine segmentation results and restore original images with these refined segmentation. 

Our proposed method contains two complementary components: Semantically-Guided Adaptation, which exploits semantic information from degraded images to refine the inaccurate segmentation results; and Exemplar-Guided Synthesis, which synthesizes restored images from semantic label maps given  degraded exemplars.

{\bf Semantically-Guided Adaptation} aims to produce `refined' segmentation maps of degraded images.
That is, this component takes a degraded image $I_d$ and its corresponding inaccurate segmentation result $S_d$ as input and produces the `refined' segmentation result. 
Taking another look at this refinement process, we can formulate it as the adaption of segmentation from degraded results to favorable ones. 
For this purpose, we design a refinement network. It learns the adaption from degraded results to favorable ones, exploits the difference, and leverages semantic information to help refine the segmentation results. 

{\bf Exemplar-Guided Synthesis} aims to generate restored images $I_r$ by deploying a restoration network. 
It takes the refined segmentation $S_r$ and the original degraded image $I_d$  as input, and outputs a recovered image $I_r$. Label-based image synthesis is a typical one-to-many translation problem~\citep{Cordts2016Cityscapes,Cordts2015Cvprw}, thus we use the original degraded image as external exemplar to control the global appearances of the output image. That is, the restored image is reconstructed from its segmentation while respecting the constraints specified by the given guidance image.
This recovery process can be formulated as guided image synthesis.

Without the loss of generality, the dataset can be defined as $\mathcal{D}=\{(I_d^1, S_d^1, I_{gt}^1, S_{gt}^1), \cdots, (I_d^N, S_d^N, I_{gt}^N, S_{gt}^N)\}$, where $I_d^i$ is a degraded image, $S_d^i$ is its original segmentation result, and $I_{gt}^i$ and $S_{gt}^i$ are the ground truth image and segmentation, respectively.

\subsection{Semantically-Guided Adaptation (SGAda)}
\label{sec:SGAda}

The refinement network aims to improve the segmentation result of the input degraded image. The adaptation from original segmentation map $S_d$ to the refined one $S_r$ can be denoted as ${\bm{G_1}: \{S_d,I_d\}\to S_r}$.

The operator $\{\cdot,\cdot\}$ means the information fusion that incorporates the source image and the guidance image.
The degraded image $I_d$ here acts as the guidance.
Typically, a guidance image reflects the desired visual effects or specified constraints and provides additional information from other modalities.
Some sophisticated generalized conditioning schemes, such as CIN~\citep{dumoulinSK2017style}, AdaIN~\citep{HuangB2017adain} and bFT~\citep{alBahar2019guided}, can be used for this process. Here we simply concatenate two inputs of SGAda and EGSyn in our experiments.

The total loss $\mathcal{L}_{\bm{G_1}}$ of the refinement network combines an adversarial loss $\mathcal{L}_{GAN_{\bm{G_1}}}$ and a refinement loss $\mathcal{L}_{\text{ref}}$ as
\begin{equation}
\mathcal{L}_{\bm{G_1}}=\mathcal{L}_{GAN_{\bm{G_1}}}+\lambda \mathcal{L}_{\text{ref}},
\end{equation}
where $\lambda$ is regularization parameters controlling the importance of two terms. 

The refinement loss $\mathcal{L}_{\text{ref}}$ is defined as a standard cross-entropy loss:
\begin{equation}
\mathcal{L}_{\text{ref}}=\mathbb{E}\left[\text{CE}\left(S_r, S_{gt}\right)\right].
\end{equation}

The adversarial loss $\mathcal{L}_{GAN_{\bm{G_1}}}$ is defined as
\begin{equation}
\begin{aligned}
\mathcal{L}_{GAN_{\bm{G_1}}} &=\mathbb{E}[\log {\bm{D_1}}(\{S_{gt}, I_d\})]\\
&+\mathbb{E}[\log (1 - {\bm{D_1}}(\{S_r, I_d\})],
\end{aligned}
\label{eq:d1}
\end{equation}
where the discriminator $\bm{D_1}$ is trained to distinguish between ground truth segmentation $S_{gt}$ and refined segmentation $S_r$ conditioned on degraded images $I_d$. $\bm{D_1}$ is formulated as an SN-PatchGAN~\citep{yu2018free} in favor of its awareness of spatial contextual relations.

\subsection{Exemplar-Guided Synthesis (EGSyn)}
\label{sec:EGSyn}
The refined segmentation result $S_r$ and original degraded image $I_d$ are then fed into the restoration network to generate restored image $I_r$. This translation can be defined as ${{\bm{G_2}}: \{I_d, S_r\} \to I_r}$.
The refined segmentation $S_r$ helps to restore the original degraded image $I_d$, while sharing the same resolution with the input image.

We train this image synthesis network with a joint loss, which consists of five terms: adversarial loss $\mathcal{L}_{GAN_{\bm{G_2}}}$, reconstruction loss $\mathcal{L}_{\ell_{1}}$, perceptual loss $\mathcal{L}_{\text{percep}}$, style loss $\mathcal{L}_{\text{style}}$ and total variation loss $\mathcal{L}_{\text{tv}}$:
\begin{equation}
\mathcal{L}_{\bm{G_2}}=\lambda_{1} \mathcal{L}_{\ell_{1}}+\lambda_{2} \mathcal{L}_{GAN_{\bm{G_2}}}+\lambda_{3} \mathcal{L}_{\text{perc}}+\lambda_{4} \mathcal{L}_{\text{style}}+\mathcal{L}_{\text{tv}}.
\end{equation}

The adversarial loss is defined similarly to (\ref{eq:d1}), as
\begin{equation}
\begin{aligned}
\mathcal{L}_{GAN_{\bm{G_2}}} = \mathbb{E}[\log \bm{D_2}(I_{gt})]  +\mathbb{E}\log [1 - \bm{D_2}(I_r)].
\end{aligned}
\end{equation}

The reconstruction loss $\mathcal{L}_{\ell_{1}}$ models the differences between reference and generated images:
\begin{equation}
\mathcal{L}_{\ell_{1}} = \mathbb{E} [\|I_{gt} - I_r\|_{1}].
\end{equation}

The perceptual loss is proposed by~\citet{Johnson2016Perceptual} based on perceptual similarity. It is originally defined as the distance between two activated features of a pre-trained deep neural network. Here we adopt a more effective perceptual loss which uses features before activation layers~\citep{wang2018esrgan}. These features are more dense and thus provide relatively stronger supervision, leading to better performance:
\begin{equation}
\mathcal{L}_{\text {perc}}=\mathbb{E}[\|\phi_{i}(I_{gt})-\phi_{i}(I_r)\|_{1}],
\end{equation}
where $\phi_{i}$ denotes the feature maps before activation of the VGG-19 network pre-trained for image classification. The employed perceptual loss encourages the model to learn how to use the extracted content information to do the translation while preserving semantic consistency.

The style loss is adopted in the same form as in the original work~\citep{Johnson2016Perceptual, Gatys2016transfer} , which aims to measure differences between covariances of activation features:
\begin{equation}
\mathcal{L}_{\text {style}}=\mathbb{E}_{j}\left[\left\|G_{j}^{\phi}\left(I_{gt}\right)-G_{j}^{\phi}\left(I_r\right)\right\|_{1}\right],
\end{equation}
where $G_{j}^{\phi}$  represents the gram matrix constructed from feature maps $\phi_{j}$. It has been shown to be an effective tool to alleviate `checkerboard' artifacts caused by transpose convolution layers~\citep{Sajjadi2017EnhanceNet}.

We also add the total variation loss to remove unwanted noises and encourage spatial smoothness in the generated images. It is useful  to mitigate checkerboard artifacts from the perceptual loss term. It is defined on the basis of the absolute gradient of generated images:
\begin{equation}
  \mathcal{L}_{\text{tv}}=\left\|{\nabla_x I_r}-{\nabla_y I_r}\right\|_{1}.
\end{equation}

For stable training, high image quality and considerable diversity, we use the least-squares GAN~\citep{mao2017least} in our experiment.

\subsection{Cooperative Modification Learning Algorithm}
\label{sec:Training strategy}
Our model consists of two components: SGAda and EGSyn. SGAda aims to exploit the difference of distributions between degradation and degradation-free results and adapt the segmentation. EGSyn tries to reconstruct the image from the segmentation and leverage the original degraded image as exemplar. To get an optimal model, we design a training strategy named \textit{Cooperative Modification Learning Algorithm}. More specifically, the training processes are divided into three stages. Firstly, we train the SGAda module including $\bm{G_1}$ and $\bm{D_1}$, using the ground truth label as supervision. Meanwhile, we train the EGSyn module including $\bm{G_2}$ and $\bm{D_2}$, using the refined map together with the original degraded image as input and the ground truth images as supervision. We then jointly train both SGAda and EGSyn in an end-to-end way until convergence.

The idea behind this strategy is that the quality of image restoration results mainly depends on the refined segmentation produced by SGAda. Although EGSyn has the degradation as exemplar, the layout of restoration mainly depends on the generated segmentation and the degradation exemplar decides other elements like textures. 
The EGSyn back-propagates the errors to guide the training to the direction that learns more precise segmentation results. Both components promote each other.

The training strategy is summarized in Algorithm~\ref{alg:training strategy}. Forward and backward in Algorithm~\ref{alg:training strategy} represent forward propagation and back propagation, respectively. The forward process includes steps of passing the input through the network layers and calculating the actual output and losses of the model. The backward process back-propagates errors and updates weights of the network. We refer corresponding operations to as forward and backward for simplicity and emphasize that our method is an end-to-end method with three-stage training. The $N_1, N_2, N_3$ are iteration numbers large enough to guarantee convergence.

\begin{algorithm}[t]
\DontPrintSemicolon
\textbf{Stage 1: SGAda Training}\\
\KwIn{$\mathbf{I_d}$ and $\mathbf{S_d}$, degraded image and its segmentation.}
\KwOut{$\mathbf{S_r}$, refined segmentation.}

\While{$n \leq N_1$}{
$\mathbf{S_r}$, $\mathcal{L}_{\bm{G_1}}$, $\mathcal{L}_{\bm{D_1}}$ = $\bm{G_1}$.forward$(\{S_d, I_d\})$\;
$\bm{G_1}$.backward
}

\textbf{Stage 2: EGSyn Training}\\
\KwIn{$\mathbf{I_d}$ and $\mathbf{S_r}$, degraded image and its refined segmentation.}
\KwOut{$\mathbf{I_r}$, restored image.}

\While{$n \leq N_2$}{
 $\mathbf{I_r}$, $\mathcal{L}_{\bm{G_2}}$, $\mathcal{L}_{\bm{D_2}}$ =  $\bm{G_2}$.forward$(\{S_{gt}, I_d\})$\;
$\bm{G_2}$.backward
}
\textbf{Stage 3: Joint Training}\\
\KwIn{$\mathbf{I_d}$ and $\mathbf{S_d}$, degraded image and its segmentation.}
\KwOut{$\mathbf{I_r}$, restored image.}

\While{$n \leq N_3$}{
 $\mathbf{S_r}$, $\mathcal{L}_{\bm{G_1}}$, $\mathcal{L}_{\bm{D_1}}$ = $\bm{G_1}$.forward$(\{S_d, I_d\})$\;
 $\mathbf{I_r}$, $\mathcal{L}_{\bm{G_2}}$, $\mathcal{L}_{\bm{D_2}}$ = $\bm{G_2}$.forward$(\{S_r, I_d\})$\;
$\bm{G_1}$.backward\\
$\bm{G_2}$.backward
}
\caption{{\sc Cooperative Modification Learning Algorithm}}
\label{alg:training strategy}
\end{algorithm}

\section{Experiments}
\label{sec:experiment}

\subsection{Experiment Settings}
We conduct experiments on different degradation types and various datasets. 
Specifically, The degradation types are divided into two categories: one is the {\bf Regular Degradation Types}, which contains the four common degradations including noise, blur, JPEG compression and chromatic aberrations; the other is the {\bf Adverse Cityscapes} under foggy, rainny and reflection conditions, respectively. With these two categories of degradations, we expect to validate the applicability of our method in both general scenes and a specific application -- autonomous driving.

We use the pre-trained models for normal conditions to compute the original segmentations, i.e., DeepLabv3~\cite{Chen2017deeplabv3} on the Cityscapes dataset~\citep{Cordts2016Cityscapes}, ResNet50dilated\footnote{https://github.com/CSAILVision/semantic-segmentation-pytorch} on the MIT ADE20K dataset,  DeepLabV2~\citep{Chen2018DeepLabv2}\footnote{https://github.com/kazuto1011/deeplab-pytorch} on the COCO-Stuff / PASCAL VOC 2012 dataset, and BiSeNet~\citep{Yu2018BiSeNet}\footnote{https://github.com/zllrunning/face-parsing.PyTorch} on the CelebAMask-HQ dataset.

For semantic segmentation, we use popular Pixel Accuracy (PA), mean Pixel Accuracy (mPA), mean Intersection over Union (mIoU), Frequency Weighted Intersection over Union (FWIoU) as evaluation metrics. For image restoration, we use PSNR and SSIM as evaluation metrics.

\subsubsection{Regular Degradation Types and Datasets}
The regular degradation types contain six common degradations including noise, blur, JPEG compression and chromatic aberrations. These degradations can be caused by moving platforms or transmission procedure, and are common in emerging applications like autonomous driving.

\begin{itemize}
\item \textit{Gaussian Blur.}  2D circularly symmetric Gaussian blur kernels are applied with standard deviations set to be [1.2, 2.5, 6.5, 15.2].
\item \textit{Gaussian Noise.} The local variance of the gaussian noise added is set to be [0.05, 0.09,0.13, 0.2].
\item \textit{JPEG Compression.} The quality factor that determines the DCT quantization matrix is set to be [43, 12,7, 4].
\item \textit{Chromatic Aberrations.} Color images are often degraded by the residual chromatic aberrations of the optical system, causing chromatism errors and a decrease of resolution. For chromatic aberration recovery, the mutual shifts of R and B channels are set to be [2, 6, 10, 14] and [1, 3, 5, 7], respectively.
\end{itemize}

We obtain the degraded images using the following datasets, which contain diverse scenes.

\begin{itemize}
\item \textit{CelebAMask-HQ}~\citep{CelebAMask-HQ} is a large-scale face image dataset that has 30,000 high-resolution face images selected from the CelebA dataset~\citep{LiuLWT15CelebA} by following CelebA-HQ~\citep{KarrasALL18CelebA-HQ}. Each image has segmentation mask of facial attributes corresponding to CelebA. The masks of CelebAMask-HQ were manually-annotated with the size of $512 \times 512$ and 19 classes including all facial components and accessories such as skin, nose, eyes, eyebrows, ears, mouth, lip, hair, hat, eyeglass, earring, necklace, neck, and cloth.
\item \textit{COCO-Stuff}~\citep{Caesar2018COCO-Stuff} is derived from the COCO dataset~\cite{Lin2014COCO}. It has 118,000 training images and 5,000 validation images captured from diverse scenes with 182 semantic classes.
\item \textit{PASCAL VOC 2012}~\citep{Everingham2010VOC} contains nearly 10K images annotated with pixel-wise segmentation of each object present.
\item \textit{ADE20K}~\citep{zhou2017scene} contains more than 20K scene-centric images exhaustively annotated with objects and object parts. Specifically, the benchmark is divided into 20K images for training, 2K images for validation, and another batch of held-out images for testing. There are totally 150 semantic categories included for evaluation, which include stuff like sky, road, grass, and discrete objects like person, car and bed.
\end{itemize}

\subsubsection{Adverse Cityscapes and Datasets}
\begin{figure}[ht]
  \centering
  \centerline {\includegraphics[width=\linewidth]{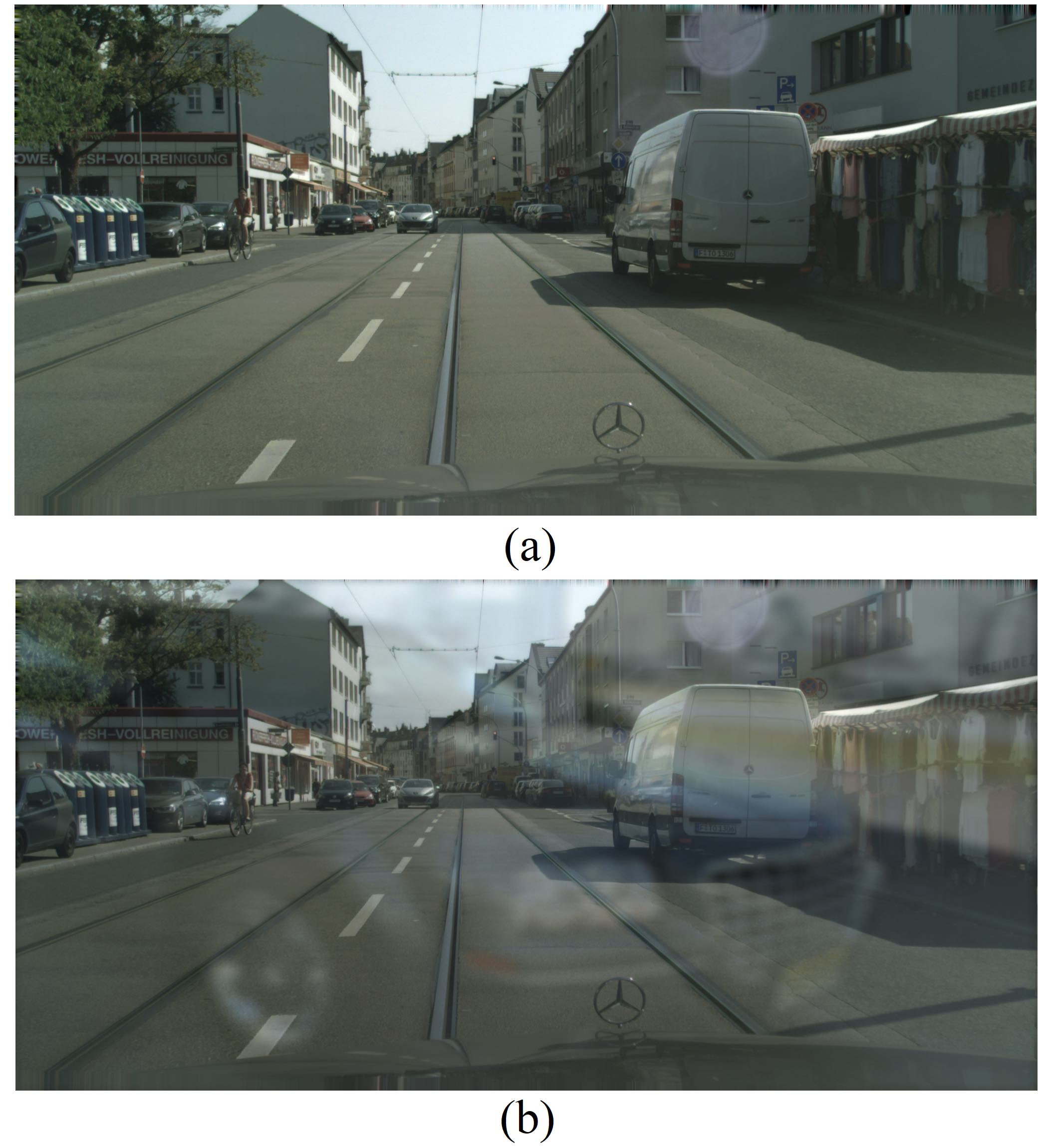}}
  \caption{Example image from Cityscapes and the result of our reflection simulation. (a) Input image from Cityscapes, (b) Output of our reflection simulation.}
  \label{fig:reflection_simulation}
\end{figure}

Bad weather is frequent, its presence could greatly affect the visibility of objects and scene in the captured photos. The Adverse Cityscapes Datasets drive from the original Cityscapes~\citep{Cordts2016Cityscapes}, which are especially proposed for autonomous driving in adverse weather and illumination conditions. 
For now, the official Cityscapes website\footnote{www.cityscapes-dataset.com} contains foggy and rainy images. 
We experiment on these two common weather phenomena and an extra image corruption source, i.e., window reflection. Reflection is a frequently-encountered source of image corruption that can arise when shooting through a glass surface. We analyze the physical properties of reflection and formulate a reflection imaging process. By then, we prepare a new dataset for reflection removal. 
\begin{itemize}
\item \textit{Fog}. 
The dataset of ~\citet{Dai2018IJCV} drives from the Cityscapes and constitutes a collection of synthetic foggy images generated with a novel fog simulation method that automatically inherit the semantic annotations of their real, clear counterparts. This dataset, named `FoggyCityscapes', contains 5000 images for training, validation and testing.
\item \textit{Rain}. 
The dataset of~\citet{Hu2019CVPR} is a new dataset with rain and fog based on the formulation of rain images. The images are adopted by using the camera parameters and scene depth information in Cityscapes to synthesize rain and fog on the photos. This dataset, named `RainCityscapes', has 9,432 training images and 1,188 test images.
\item \textit{Reflection}. 
For image synthesis, we use two methods to synthesis reflection images as in~\citep{wei2019single, zhang2018single, fan2017generic} with some modifications to make sure the size of obtained images is identical with those in~\citep{Cordts2016Cityscapes}. The images from~\citep{Cordts2016Cityscapes} and~\citep{Everingham2010VOC} are used as synthetic and reflective images, respectively. We name this dataset `ReflectCityscapes' after the Cityscapes. Results of the presented pipeline for reflection simulation on example images from Cityscapes are provided in Figure~\ref{fig:reflection_simulation}.
More samples and the reflection simulation scripts can be found at~\href{xiaweihao.com/projects/sr-net}{our website}. 
\end{itemize}

\begin{figure}[ht]
  \centering
  \centerline {\includegraphics[width=\linewidth]{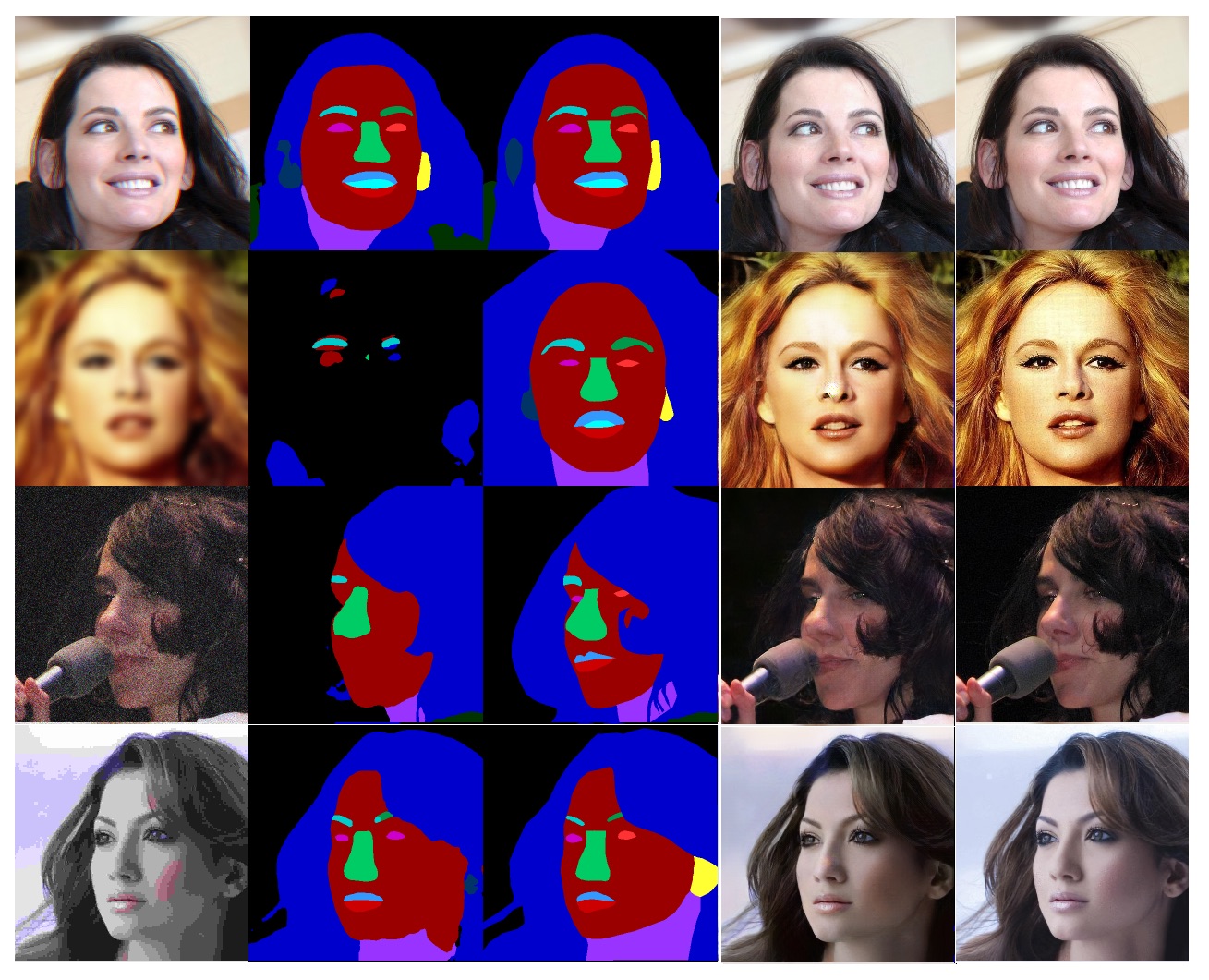}}
  \caption{Qualitative results of semantic segmentation and image restoration on \textit{CelebAMask-HQ}~\citep{CelebAMask-HQ}. From left to right: (a) Input image, (b) Original segmentation result of the input image using~\citep{Yu2018BiSeNet}, (c) Refined segmentation, (d) Restored image without any post-processing, (e) Ground truth image. The degradation types from top to bottom are CA, GB, GN and JPEG, respectively.}
  \label{fig:celeba_results}
\end{figure}

\subsection{Results of Regular Degradation Types}
\label{sec:results}
To validate the applicability of our proposed method for {\bf Regular Degradation Types}, we conduct evaluations on various datasets including \textit{CelebAMask-HQ}~\citep{CelebAMask-HQ}, \textit{COCO-Stuff}~\citep{Caesar2018COCO-Stuff}, \textit{VOC 2012}~\cite{Everingham2010VOC} and \textit{ADE20K}~\citep{zhou2017scene}. 
Section~\ref{sec:seg_regular} presents the semantic segmentation results of certain degradations with their corresponding ground truth segmentations.
Section~\ref{sec:res_regular} is for the comparison with state-of-the-art image restoration methods proposed for multiple degradations.

\begin{table*}[ht]
\caption{Quantitative results on \textit{CelebAMask-HQ}~\citep{CelebAMask-HQ}. CA, GB, GN and JPEG indicate Chromatic Aberrations, Gaussian Blur, Gaussian Noise and JPEG Compression, respectively. GB\_Re means Refinement or Restoration results of Gaussian Blur images. For semantic segmentation, we use Pixel Accuracy (PA), mean Pixel Accuracy (mPA), mean Intersection over Union (mIoU), Frequency Weighted Intersection over Union (FWIoU) as evaluation metrics. For image restoration, we use PSNR and SSIM as evaluation metrics.
}
\label{tab:quantitative results}
\begin{center}
\begin{tabular}{c|cccc|cc}
\toprule
\multirow{2}{*}{Degradation} &\multicolumn{4}{c|}{Semantic Segmentation}&\multicolumn{2}{c}{Image Restoration}\\
\cline{2-7}
 & PA	&mPA	&mIoU &	FWIoU  &PSNR & SSIM \\
\hline\hline

Original        &0.9598	&0.8725	 &0.8049	&0.9268	&-	&-\\
\hline
CA	             &0.9361	&0.7332	&0.6453	&0.8893	&23.05 &0.7407\\
CA\_Re	     &0.9569	&0.8531	&0.7150	&0.9274  &32.03 &0.9271\\	\hline

GB                &0.4023	&0.2289  &0.1524	&0.2265  &23.93 &0.6960\\
GB\_Re       &0.5808 &0.4554  &0.4156  &0.5916  &30.97 &0.9331\\	\hline

GN	             &0.9402	&0.7720	&0.6923	&0.9002  &22.32 &0.5946\\
GN\_Re	     &09664	&0.8530	&0.7505	&0.9172  &28.37 &0.8388\\\hline

JPEG	         &0.9224	&0.7237	&0.6311	&0.8676  &25.67 &0.7482\\
JPEG\_Re	 &0.9280 &0.7947 &0.7139  &0.8863  &27.73 &0.8384\\
\bottomrule
\end{tabular}
\end{center}
\end{table*}

\subsubsection{Semantic Segmentation of Regular Degradation Types}
\label{sec:seg_regular}
The first three columns of Figure~\ref{fig:celeba_results} show the input degraded images, original segmentation results computed by~\citep{Yu2018BiSeNet} and our segmentation.
For visualization purposes, we follow the color encoding scheme of \textit{CelebAMask-HQ}~\citep{CelebAMask-HQ} to colorize the label map.
As shown, the obtained semantic maps of images degraded by a specific degradation are terrible since the segmentation method we adopted are designed to work under optimal conditions.

The quantitative comparison results are shown in Table~\ref{tab:quantitative results}. The segmentation results of original images without any degradation are presented in the first row. Performances rapidly deteriorate when the input images are corrupted with different degradation types. We quantify the extent of segmentation for different degradation types. Specifically, images corrupted by gaussian blur get the lowest segmentation performance. Other degradation types have relatively less influence, which is also in accordance with the observation in Figure~\ref{fig:celeba_results}.

For corrupted images of different degradation types, the refinement network improves the segmentation performance with a large margin. The experiments demonstrate the universality and scalability of our proposed method. In the case of gaussian noise, the performance even outperforms the clear images.
As illustrated, our approach leads to satisfactory results with virtually no artifacts.

\begin{figure*}[ht]
  \centering
  \centerline {\includegraphics[width=\textwidth]{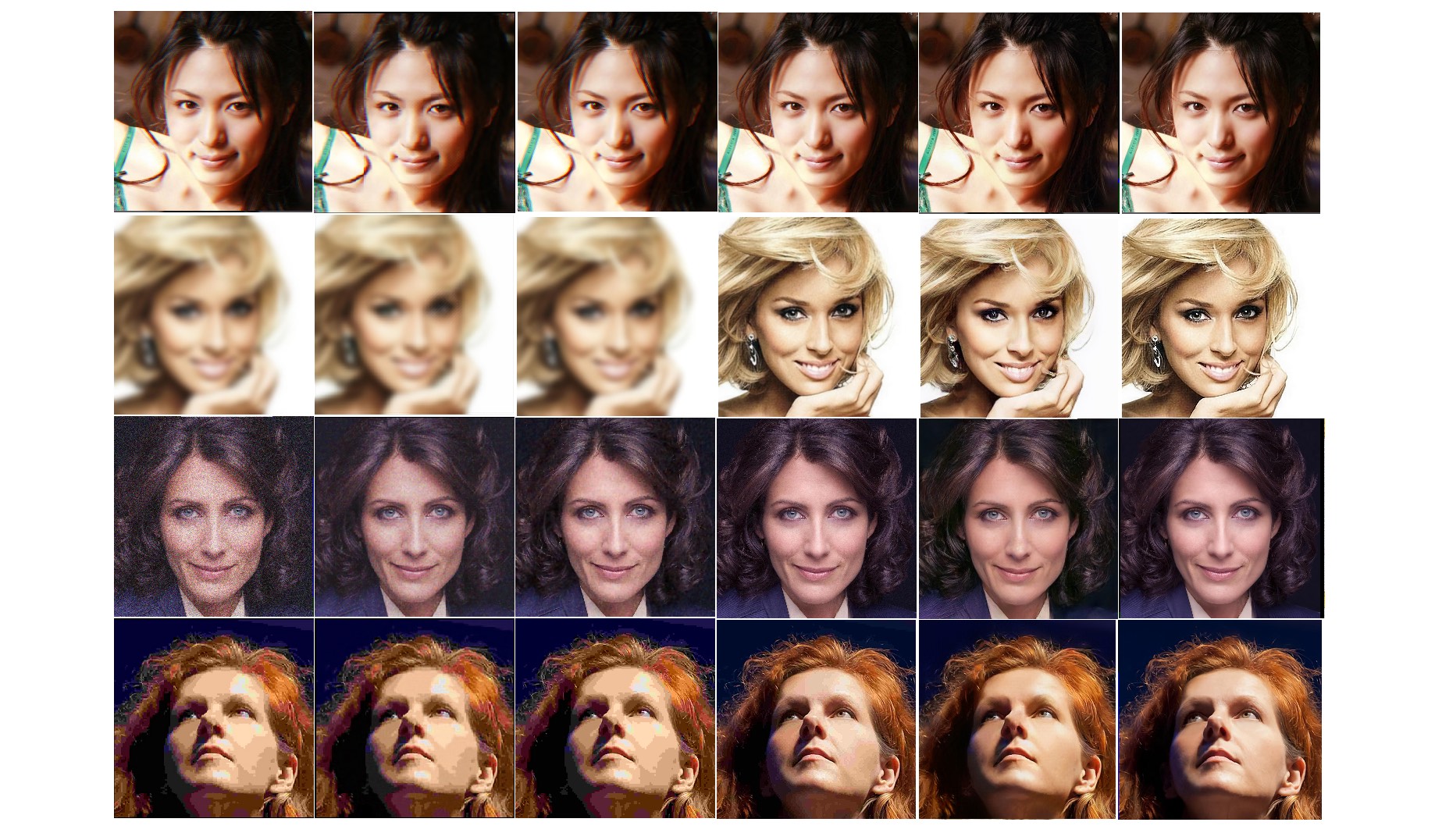}}
  \caption{Qualitative comparison of state-of-the-art image restoration methods on \textit{CelebAMask-HQ}~\cite{CelebAMask-HQ}. From left to right: (a) Degradation, (b) Deep Prior~\citep{Ulyanov2018Prior}, (c) RL-Restore~\citep{Yu2018RLRestore}, (d) NLRN~\citep{Liu2018NLRN} (e) our proposed method, and (f) ground truth. Better seen on screen and zoomed in.}
  \label{fig:comparison_results}
\end{figure*}

\begin{table*}[ht]
\caption{Quantitative results of different degradation types on \textit{CelebAMask-HQ}~\citep{CelebAMask-HQ}. CA, GB, GN and JPEG denote Chromatic Aberrations, Gaussian Blur, Gaussian Noise and JPEG Compression, respectively. The {\bf best} performance is in bold.}
\label{tab:quantitative comparison}
\begin{center}
\begin{tabular}{c|cc|cc|cc|cc|cc}
\toprule
\multirow{2}{*}{Degradation} &\multicolumn{2}{c|}{Degraded}&\multicolumn{2}{c|}{Deep Prior~\citep{Ulyanov2018Prior}} &\multicolumn{2}{c|}{RL-Restore~\citep{Yu2018RLRestore}}&\multicolumn{2}{c|}{NLRN~\citep{Liu2018NLRN}}&\multicolumn{2}{c}{Ours}\\
\cline{2-11}
 &PSNR & SSIM &PSNR & SSIM &PSNR & SSIM &PSNR & SSIM&PSNR & SSIM\\
\hline\hline	
CA	    &23.05  &0.7407  &23.25  &0.7539 &31.17 &0.9025 &31.83 &0.9214 &{\bf 32.03} &{\bf 0.9271}\\
GB	    &24.11  &0.7562  &24.58  &0.7608 &28.40 &0.8827 &29.71 &0.8864 &{\bf 30.97} &{\bf 0.8931}\\	
GN	    &22.32  &0.5946  &22.45  &0.6167 &26.36 &0.8011 &28.11 &0.8339 &{\bf 28.37} &{\bf 0.8388}\\
JPEG	&25.67  &0.7482  &26.11  &0.7523 &27.09 &0.8115 &27.52 &0.8295 &{\bf 27.73} &{\bf 0.8384}\\					
\bottomrule
\end{tabular}
\end{center}
\end{table*}

\subsubsection{Image Restoration of Regular Degradation Types}
\label{sec:res_regular}
We further compare our method with state-of-the-art image restoration approaches proposed for multiple degradations. Deep Prior~\citep{Ulyanov2018Prior} is non-trained image restoration method for image denosing and super-resolution. RL-Restore~\citep{Yu2018RLRestore} adopts an agent to select a toolchain to progressively restore an image corrupted by complex and mixed distortions. DnCNN~\citep{ZhangZCM017DnCNN}, VDSR~\citep{Kim2016VDSR}, MemNet~\citep{TaiYLX17MemNet} and NLRN~\citep{Liu2018NLRN} are the state-of-the-art learning-based models for handling multiple degradations. Since RL-Restore and NLRN have claimed their superiority over DnCNN, VDSR and MemNet, we didn't compare with these methods.

The results of all methods are summarized in Table~\ref{tab:quantitative comparison}. Bold indicates the best performance. Visual comparison are illustrated in Figure~\ref{fig:comparison_results}. The restoration results of Deep Prior~\citep{Ulyanov2018Prior} are not good, which perhaps can be due to the sensitivity of hyper-parameters that we didn't deliberately select for each degraded image. In most cases of RL-Restore~\citep{Yu2018RLRestore}, the results are quite satisfactory. However, the performance drops, sometimes image barely ameliorates when the distortions are extremely severe. NLRN~\citep{Liu2018NLRN} performs good on image denoising and image dejpeg as in the original paper, we surprisingly found that it also produce comparatively satisfactory results on image deblur task. Our method produces best quantitative and visual results.

From these results, we observe that the segmentation and restoration results generally get improved with our approach.
Due to the semantically-guided adaptation, semantic information from degraded images is exploited and the information is then leveraged to refine the original segmentation. Exemplar-guided synthesis learns how to generate restored images from semantic label maps together with specific degraded exemplars.

\begin{figure*}[ht]
  \centering
  \centerline {\includegraphics[width=\textwidth]{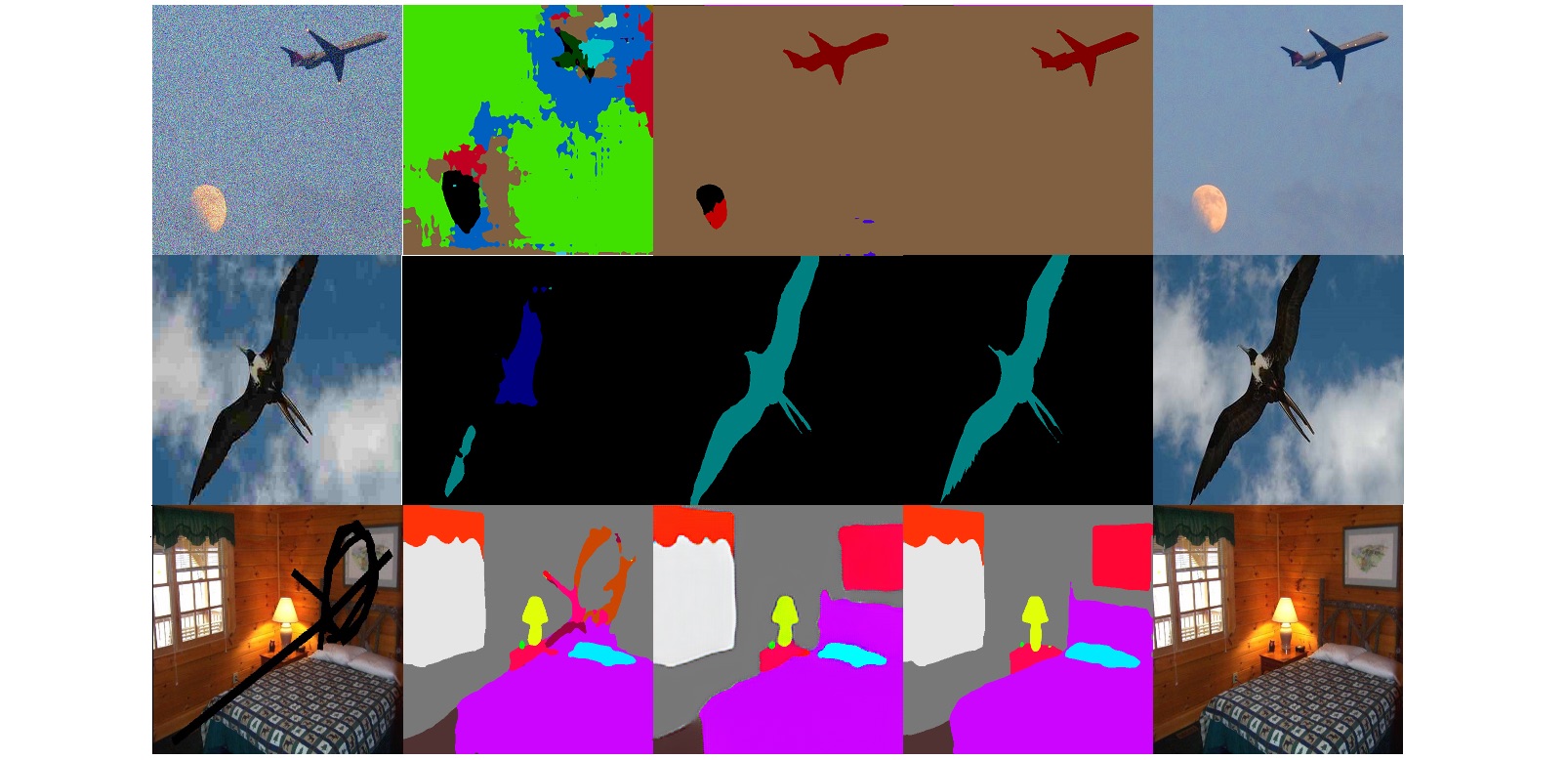}}
  \caption{Qualitative results on \textit{COCO-Stuff}~\citep{Caesar2018COCO-Stuff} (GN), \textit{PASCAL VOC 2012}~\citep{Everingham2010VOC} (JPEG), and \textit{ADE20K}~\citep{zhou2017scene} (IM). From left to right: (a) Input image, (b) Original segmentation result of the input image using~\citep{Yu2018BiSeNet} and~\citep{Chen2018DeepLabv2}, (c) Refined segmentation, (d) Ground truth segmentation, and (e) Restored image.}
  \label{fig:more_results}
\end{figure*}

Extensive qualitative results of COCO-Stuff, PASCAL VOC 2012, and ADE20K are demonstrated in Figure~\ref{fig:more_results}.

\subsection{Results of Adverse Cityscapes}
\label{sec:results on cityscapes}
To validate our method on autonomous driving scenes, we experiment on {\bf Adverse Cityscapes} (i.e., fog, rain, reflection).
We evaluate the proposed method for semantic segmentation in Section~\ref{sec:seg_cityscapes} and compare restoration results with state-of-the-art defog, derain and dereflection methods in Section~\ref{sec:res_cityscapes}.

\subsubsection{Semantic Segmentation of Adverse Cityscapes}
\label{sec:seg_cityscapes}
\begin{figure*}[th]
  \centering
  \centerline {\includegraphics[width=\textwidth]{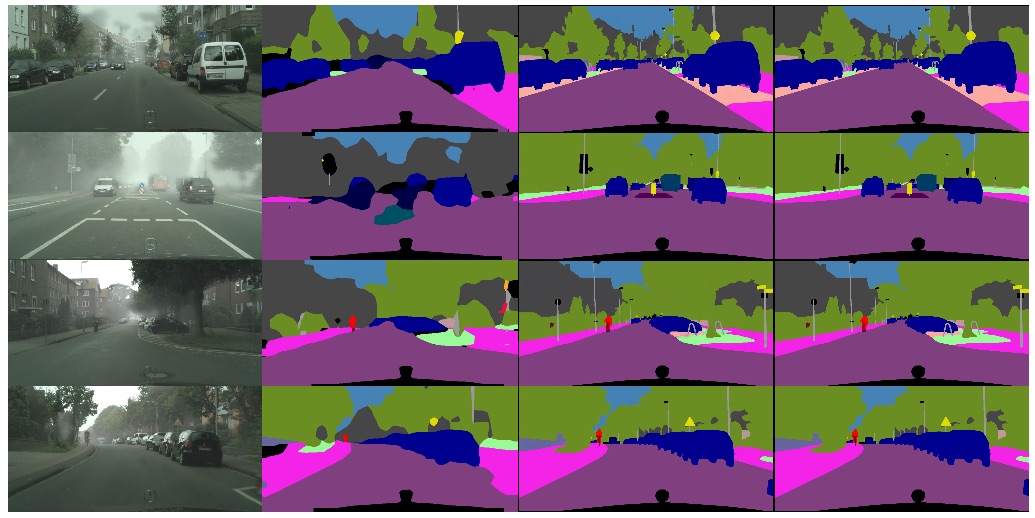}}
  \caption{Qualitative results for semantic segmentation on FoggyCityscapes~\citep{Dai2018IJCV}. From left to right: (a) Input foggy image, (b) Original segmentation of the foggy image using~\cite{Chen2017deeplabv3}, (c) Our segmentation of foggy image, (d) Ground truth segmentation of the original fog-free image.}
  \label{fig:results_fog_seg}
\end{figure*}

\begin{figure*}[th]
  \centering
  \centerline {\includegraphics[width=\textwidth]{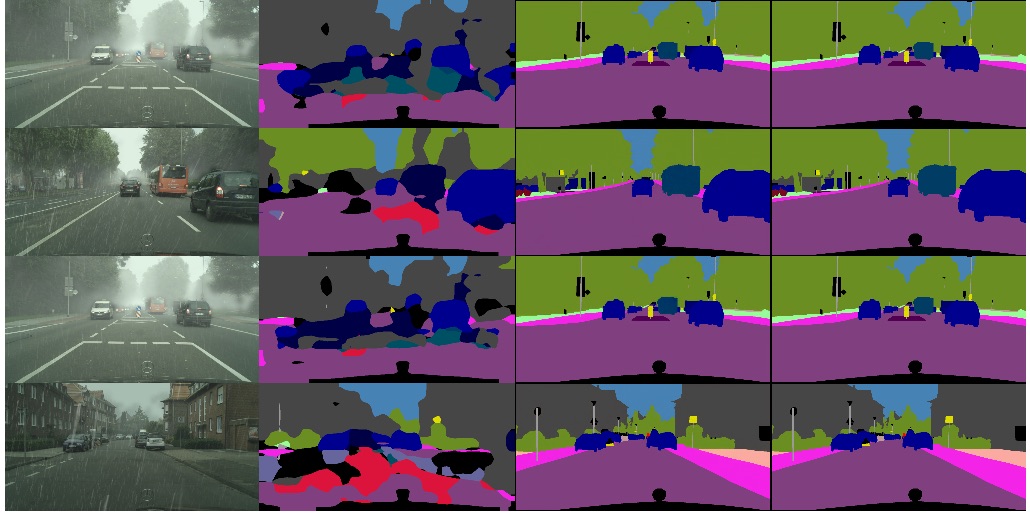}}
  \caption{Qualitative results for semantic segmentation on RainCityscapes~\citep{Hu2019CVPR}. From left to right: (a) Input rainny image, (b) Original segmentation of the rainy image using~\cite{Chen2017deeplabv3}, (c) Our segmentation of rain image, (d) Ground truth segmentation of the original rain-free image.}
  \label{fig:results_rain_seg}
\end{figure*}

\begin{figure*}[th]
  \centering
  \centerline {\includegraphics[width=\textwidth]{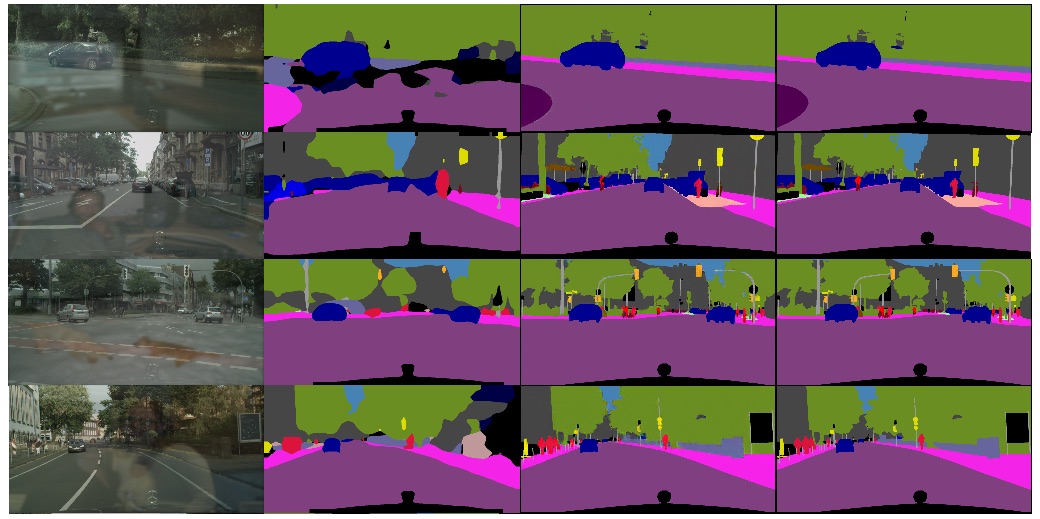}}
  \caption{Qualitative results for semantic segmentation on our ReflectCityscapes. From left to right: (a) Input reflected image, (b) Original segmentation of the reflected image using~\cite{Chen2017deeplabv3}, (c) Our segmentation of Reflection image, (d) Ground truth segmentation of the original reflection-free image.}
  \label{fig:results_reflect_seg}
\end{figure*}

The main semantic segmentation experiment showcases the effectiveness of our pipeline.
Semantic segmentation results in foggy, rainy and reflection environmental conditions are shown in Figure~\ref{fig:results_fog_seg}, Figure~\ref{fig:results_rain_seg} and Figure~\ref{fig:results_reflect_seg}, respectively.

We note that the significant performance benefit delivered by Cooperative Modification Learning Algorithm. This can be attributed to the fact that images captured in adverse conditions such as fog have large intra-domain variance as a result of poor visibility, effects of artificial lighting sources and motion blur. However, we believe that domain-adversarial approaches have the potential to be used for transferring knowledge to adverse weather domains. Our method shows considerable improvement by transferring semantic knowledge from normal environmental conditions to fog, rain and reflection.

\subsubsection{Image Restoration of Adverse Cityscapes}
\label{sec:res_cityscapes}
The performance of all compared methods on the datasets is reported in Table~\ref{tab:results_derain}, Table~\ref{tab:results_defog} and Table~\ref{tab:results_deflect}, where our method performs favorably against all the others in terms of both PSNR and SSIM.

Table~\ref{tab:results_derain} shows the comparison results with the state-of-the-art rain removal methods, i.e., DSC ~\cite{Luo2015derain}, GMMLP ~\cite{Li2016derain}, JOB~\cite{Zhu2017derain}, RESCAN~\cite{Li2018derain}, DID-MDN~\cite{Zhang2018derain}, SPANet~\cite{Wang2019CVPR}, PReNet~\cite{ren2019progressive} and DAF-Net ~\cite{Hu2019CVPR}. Table~\ref{tab:results_defog} shows the comparison results with the state-of-the-art fog removal methods, i.e., DCP~\cite{He2009prior}, NLID~\cite{Berman2016NLID} and MSCNN~\cite{ren2016mscnn}. Table~\ref{tab:results_deflect} shows the comparison results with the state-of-the-art reflection removal methods, i,e., BDN~\cite{eccv18refrmv}, LB14\cite{yu2019single}, CEILNet~\cite{fan2017generic}, PLRS~\cite{zhang2018single} and ERRNet~\cite{wei2019single}.

\begin{figure*}[th]
  \centering
  \centerline {\includegraphics[width=\textwidth]{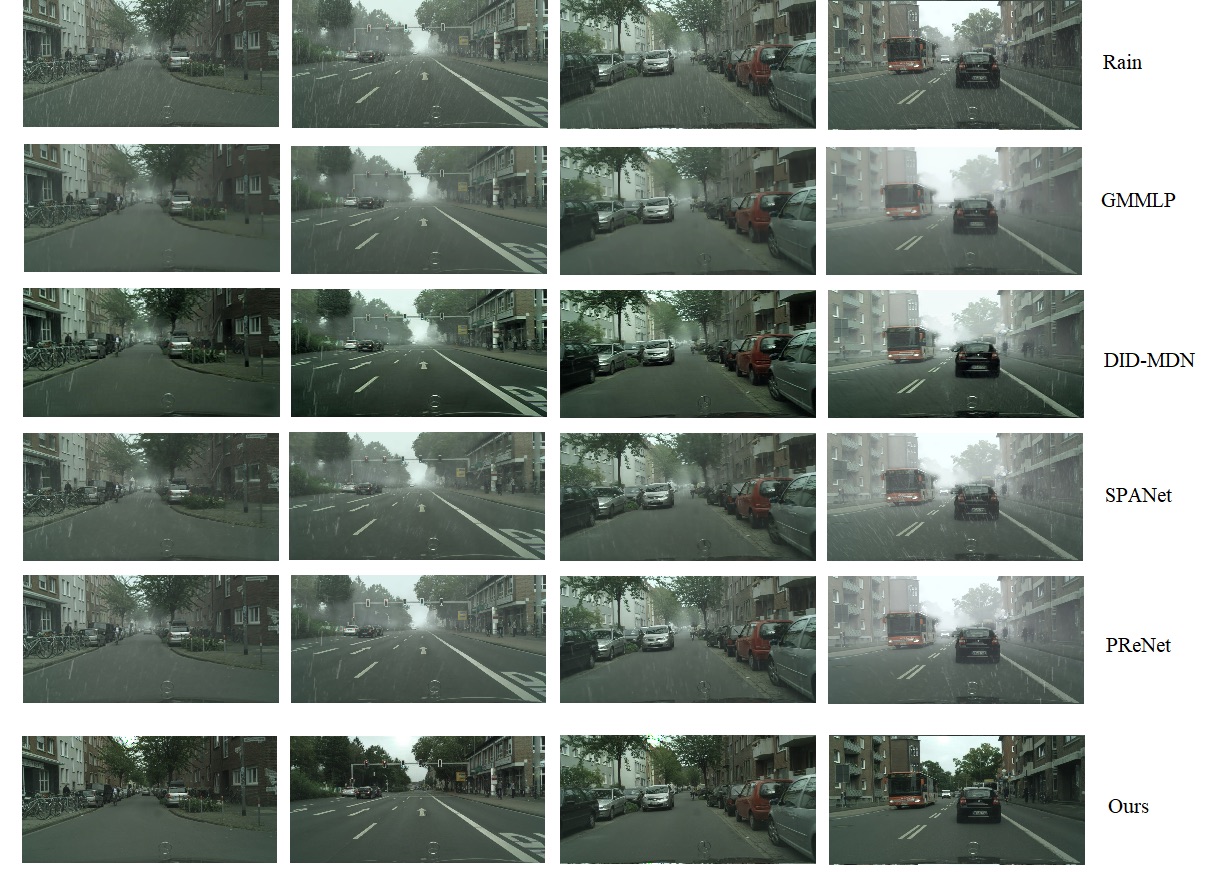}}
  \caption{Representative images from RainCityscapes and the derained versions of them obtained with the four deraining methods that we consider in our experiments on the utility of deraining preprocessing. (a) RainCityscapes image, (b) GMMLP~\cite{Li2016derain}, (c) DID-MDN\cite{Zhang2018derain}, (d) SPANet~\cite{Wang2019CVPR}, (e) PReNet~\cite{ren2019progressive}, (f) Ours. Better seen on screen and zoomed in.}
  \label{fig:results_derain}
\end{figure*}

\begin{figure*}[th]
  \centering
  \centerline {\includegraphics[width=\textwidth]{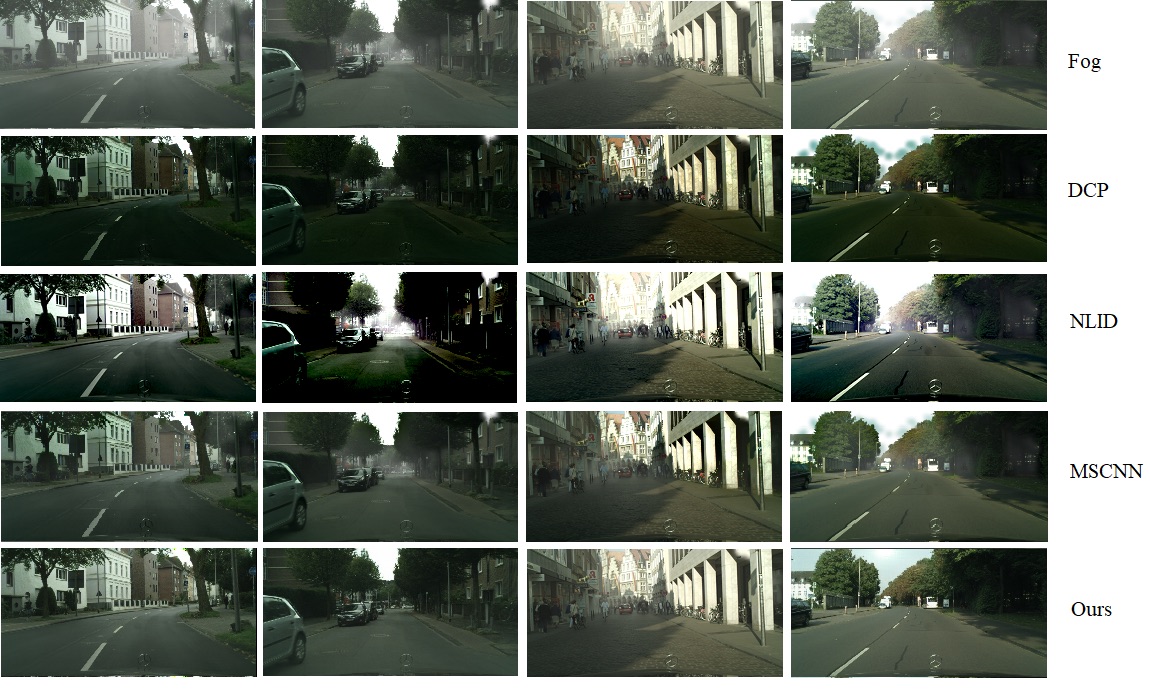}}
  \caption{Representative images from FoggyCityscapes and dehazed versions of them obtained with the three dehazing methods that we consider in our experiments on utility of dehazing preprocessing. (a) FoggyCityscapes image, (b) DCP~\cite{He2009prior}, (c) NLID~\cite{Berman2016NLID}, (d) MSCNN~\cite{ren2016mscnn}, (e) Ours. This figure is better seen on screen and zoomed in.}
  \label{fig:results_defog}
\end{figure*}

\begin{figure*}[th]
  \centering
  \centerline {\includegraphics[width=\textwidth]{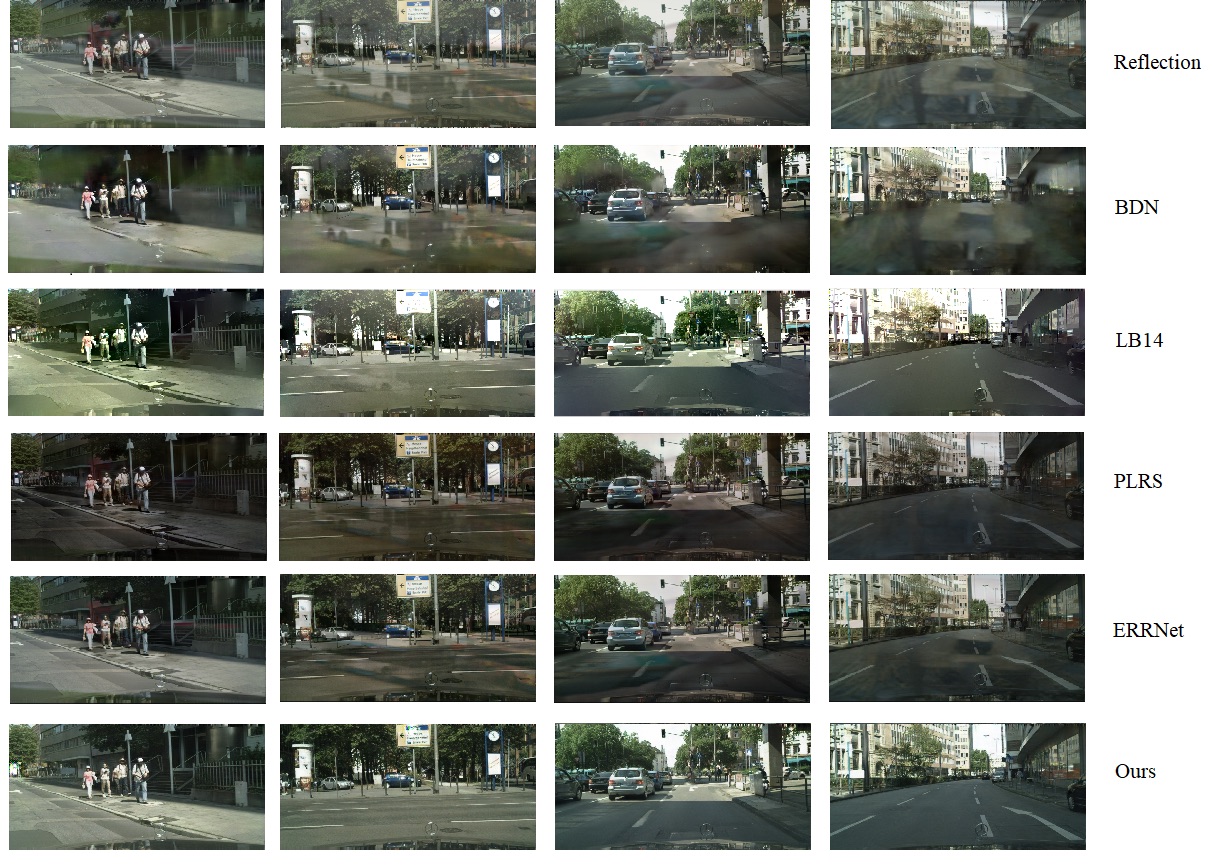}}
  \caption{Representative images from FoggyCityscapes and the dehazed versions of them obtained with the three dehazing methods that we consider in our experiments on the utility of dehazing preprocessing. (a) ReflectCityscapes image, (b) BDN~\cite{eccv18refrmv}, (c) LB14~\cite{yu2019single}, (d) PLRS~\cite{zhang2018single}, (e) ERRNet~\cite{wei2019single}, (f) Ours. Better seen on screen and zoomed in.}
  \label{fig:results_dereflect}
\end{figure*}

We illustrate the visual comparison reults of restoration for rain, fog and reflection are demonstrated in Figure~\ref{fig:results_derain}, Figure~\ref{fig:results_defog} and Figure~\ref{fig:results_dereflect}. As shown, our method can clearly remove the fog, rain streaks and reflection, while others tend to produce artifacts on the images or fail to remove large rain streaks, which are also revealed in the corresponding numerical values.

\begin{table}[t]
\centering
\caption{Comparison with the state-of-the-arts in terms of the PSNR and SSIM on the test set of RainCityscapes~\citep{Hu2019CVPR}.}
\label{tab:results_derain}
\begin{tabular}{c|c|c}
\toprule
Method& PSNR & SSIM\\
\midrule
DSC ~\cite{Luo2015derain} &16.25 &0.7746\\
GMMLP ~\cite{Li2016derain} &17.80 &0.8169\\
JOB~\cite{Zhu2017derain} &15.10 & 0.7592\\
RESCAN~\cite{Li2018derain} &24.49 &0.8852\\
DID-MDN~\cite{Zhang2018derain} &28.43 &0.9349\\
SPANet~\cite{Wang2019CVPR} &25.39&0.8933\\
PReNet~\cite{ren2019progressive} &25.96&0.9147\\
DAF-Net ~\cite{Hu2019CVPR}   &30.06 &0.9530\\
Ours &\bf{32.41} &\bf{0.9579}\\
\bottomrule
\end{tabular}
\end{table}

\begin{table}[th]
\centering
\caption{Comparison with the state-of-the-arts in terms of the PSNR and SSIM on the test set of FoggyCityscapes~\citep{Dai2018IJCV}.}
\label{tab:results_defog}
\begin{tabular}{c|c|c}
\toprule
Method& PSNR & SSIM\\
\midrule
DCP~\cite{He2009prior}  &23.98 &0.8349\\
NLID~\cite{Berman2016NLID} &24.43 &0.8512\\
MSCNN~\cite{ren2016mscnn} &29.36 &0.9317\\
Ours   &\bf{32.64} &\bf{0.9618}\\
\bottomrule
\end{tabular}
\end{table}

\begin{table}[th]
\centering
\caption{Comparison with the state-of-the-arts in terms of the PSNR and SSIM on the test set of ReflectCityscapes.}
\label{tab:results_deflect}
\begin{tabular}{c|c|c}
\toprule
Method& PSNR & SSIM\\
\midrule
LB14~\cite{yu2019single}&29.23 &0.9337\\
CEILNet~\cite{fan2017generic} &24.51 &0.8826\\
PLRS~\cite{zhang2018single} &28.06 &0.9182\\
BDN~\cite{eccv18refrmv} &15.10 & 0.7592\\
ERRNet~\cite{wei2019single} &30.80 &0.9369\\
Ours &\bf{32.06} &\bf{0.9590}\\
\bottomrule
\end{tabular}
\end{table}

\section{Conclusion}
\label{sec:conclusion}
In this paper, we have introduced a novel framework to tackle semantic segmentation and image restoration in adverse environmental conditions. 
Our proposed method achieves cooperative image restoration and semantic segmentation by combining the complementary information and inherent interdependence of the two seemingly unrelated tasks in a holistic way.
Qualitative and quantitative results demonstrate the superiority of our proposed method over the state-of-the-art methods in both semantic segmentation and image restoration.



\bibliographystyle{spbasic}      
\bibliography{reference}   

\end{document}